\pdfoutput=1
\documentclass[11pt]{article}

\usepackage[]{ACL2023}
\usepackage{amsmath}
\usepackage{times}
\usepackage{arydshln}
\usepackage{latexsym}
\usepackage{multirow}
\usepackage{graphicx}
\usepackage{colortbl}
\usepackage{array}
\usepackage{makecell}

\usepackage{tabularray}
\usepackage[T1]{fontenc}
\usepackage[utf8]{inputenc}
\usepackage{lipsum}  
\usepackage{microtype}

\usepackage{inconsolata}
\usepackage{multirow}
\usepackage{graphicx}
\usepackage{booktabs}
\usepackage{inconsolata} % For Inconsolata font
\usepackage{bbding}
\usepackage{pifont}
\usepackage{wasysym}
\usepackage{amssymb}

\usepackage[normalem]{ulem}
\useunder{\uline}{\ul}{}
\title{PicPersona-TOD : A Dataset for Personalizing Utterance Style in Task-Oriented Dialogue with Image Persona}

\author{Jihyun Lee$^1$, Yejin Jeon$^1$, Seungyeon Seo$^1$, Gary Geunbae Lee$^{1,2}$ \\
  $^1$Graduate School of Artificial Intelligence, POSTECH, Republic of Korea\\
  $^2$Department of Computer Science and Engineering, POSTECH, Republic of Korea\\
  \texttt{\{jihyunlee,  jeonyj0612, ssy319, gblee\}@postech.ac.kr} \\
}

\begin{document}
\maketitle
\begin{abstract}

Task-Oriented Dialogue (TOD) systems are designed to fulfill user requests through natural language interactions, yet existing systems often produce generic, monotonic responses that lack individuality and fail to adapt to users' personal attributes. To address this, we introduce PicPersona-TOD, a novel dataset that incorporates user images as part of the persona, enabling personalized responses tailored to user-specific factors such as age or emotional context. This is facilitated by first impressions, dialogue policy-guided prompting, and the use of external knowledge to reduce hallucinations. Human evaluations confirm that our dataset enhances user experience, with personalized responses contributing to a more engaging interaction. Additionally, we introduce a new NLG model, Pictor, which not only personalizes responses, but also demonstrates robust performance across unseen domains. \footnote{\url{https://github.com/JihyunLee1/PicPersona}}

\end{abstract}

% The contributions of this work include the development of the first PicPersona-TOD visual persona dataset, the introduction of a automatic framework for dataset generation, and comprehensive benchmarks demonstrating that enhances user experience without compromising performance in other tasks.

% https://arxiv.org/pdf/2006.05635.pdf
\section{Introduction}

Task-oriented dialogue (TOD) is one of the core tasks of dialogue systems, which is designed to fulfill user requests, such as assisting users at customer service desks \cite{sgd} and tourist centers \cite{mwoz22}. A TOD system is typically divided into the following sub-modules: (1) dialogue state tracking (DST) for tracking the user's requests, (2) policy module for determining system actions such as database (DB) searches or dialogue terminations, and (3) natural language generation module (NLG) for converting dialogue policies and DB results into natural language responses \cite{young2013pomdp}. Among these components, the responses generated by the NLG module are used to directly interact with the user; therefore, NLG responses significantly influence the overall user experience.
\begin{figure}[h]
  \centering
  \includegraphics[width=0.80\linewidth]{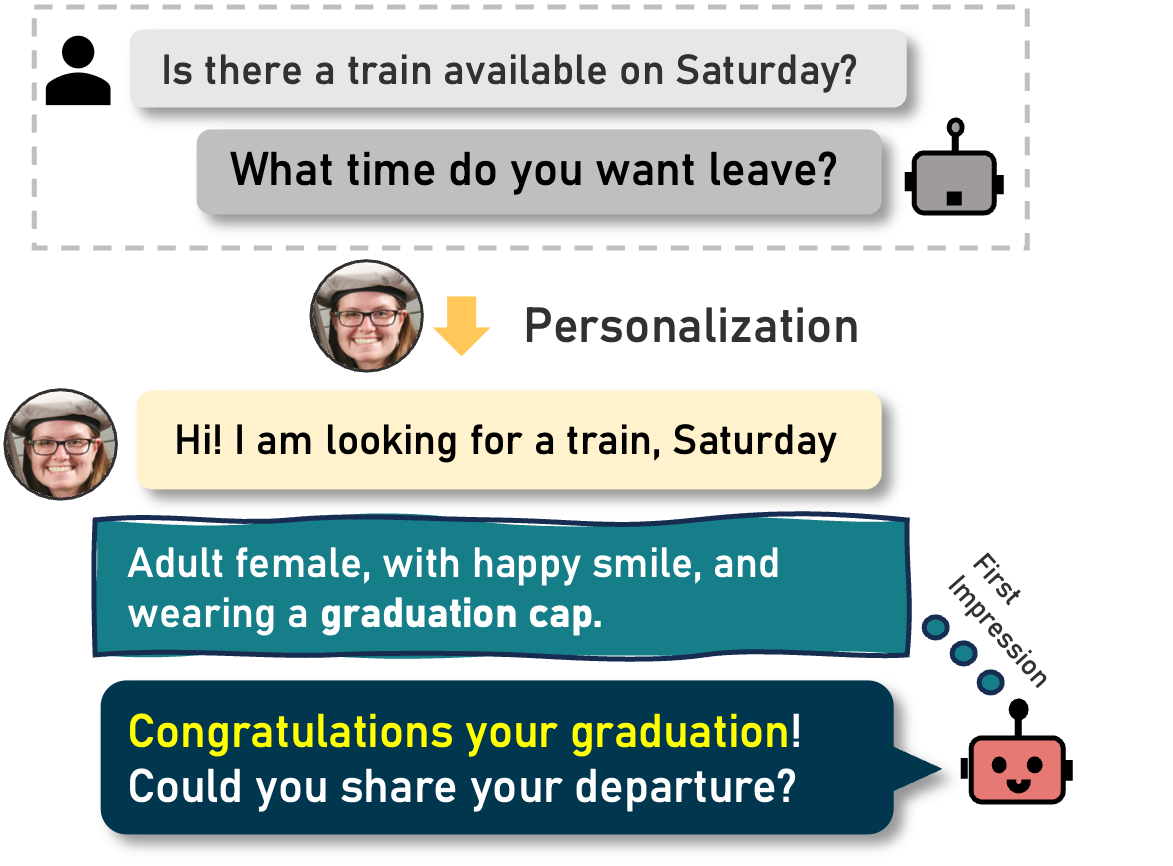}
   \caption{Example of PicPersona-TOD: Unlike existing TOD datasets (in grey), which lack user personas and personalization, PicPersona-TOD uses user images to generate tailored responses.}
   \label{fig:intro}
\end{figure}
\begin{figure*}[ht]
  \centering
  \includegraphics[width=1.0\linewidth]{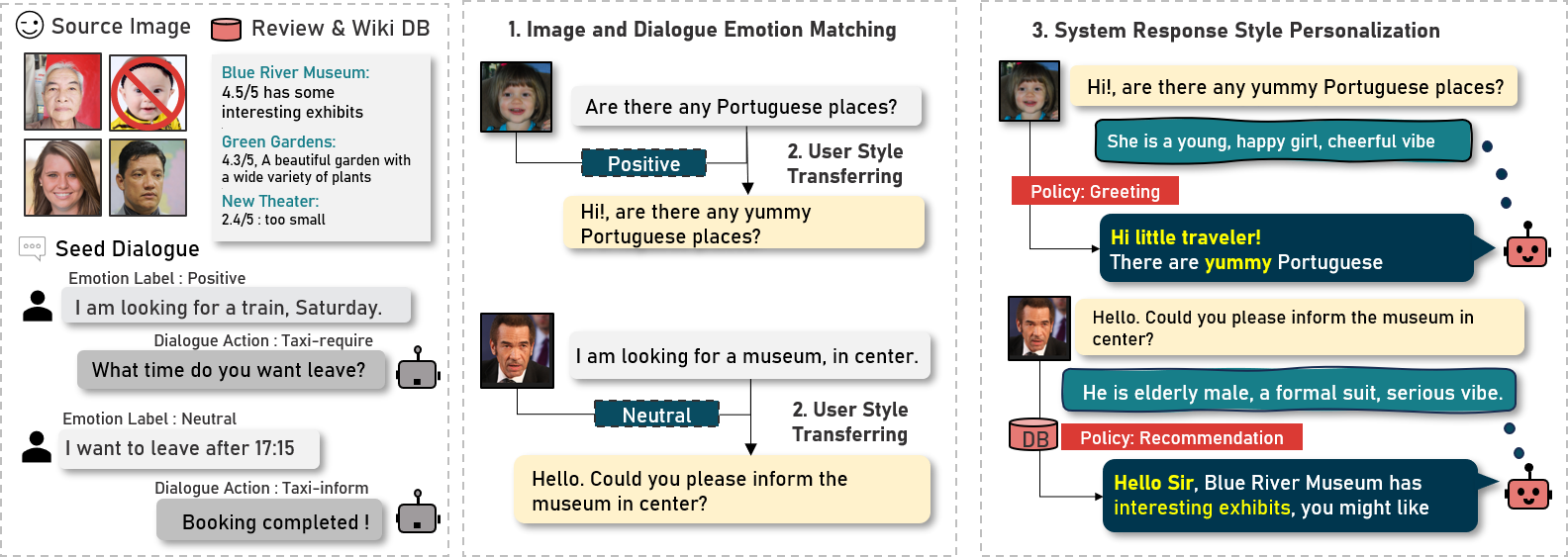}
   \caption{An overview of the automatic pipeline for generating PicPersona-TOD dataset.}
   \label{fig:main}
\end{figure*}

Although extensive research has improved system responses \cite{tod3soloist,tod2mintl,simpletod,pptod,tod1ubar,KRLS,ohashi2023enhancing}, the primary focus has remained on enhancing the accuracy of information conveyance. As a result, the style of the generated response is monotonic and lacks individuality, which hinders the system from forming age-appropriate and emotionally resonant connections with users \cite{mclean2021alexa}.  

In an effort to improve the naturalness and engagement of system responses, recent approaches have curated new TOD datasets that support personalized response styles by incorporating user personas into the dialogue. For instance, \citet{joshi2017personalization} included age and gender information in the dialogues, \citet{lin2023emous} integrated emotion, and \citet{toad} personalized system responses by mirroring users' noun and verb phrases. Although these approaches provide personalized responses to some degree, the persona modality has been limited to textual information, which lacks details and concurrency about the users they interact with. 

Meanwhile, in the field of open dialogue systems, the integration of user personas has been a long-standing point of interest \cite{persona_structured_pet, comicbook_agrawal2023multimodal,kim2024pearl,qian2017assigning, zheng2019personalized,pmg}, with recent advancements highlighting the potential of the \textit{visionary persona} approach \cite{mpchat, lee2024stark}. A visionary persona can capture subtle facial expressions and provide a rich understanding of the user's context, which is similar to how people interpret visual and non-verbal cues in real-life interactions. This approach is particularly valuable in situations where prior textual user profiles are unavailable, such as first-time interactions where no historical information is present. Despite these advantages, visionary personas have primarily been utilized in chit-chat \cite{poria2018meld, firdaus2020meisd} or counseling scenarios \cite{valstar2016avec, li2023medic}  and have not yet been explored in current TOD systems.

Taking these considerations into account, we introduce a new TOD dataset that incorporates realistic user images as part of the user persona, enabling personalized system responses in terms of greetings, formality, age sensitivity, and emotional awareness (Figure~\ref{fig:intro}). In constructing PicPersona-TOD, we use the user's first impression and dialogue policy-guided prompts, which effectively distill the personalization capabilities of Large Language Model (LLM). Additionally, we incorporate external knowledge from Google Maps and Wikipedia to reduce hallucinations in personalized responses. Furthermore, we implement a meticulous filtering process to ensure stylistic appropriateness, semantic accuracy, and overall naturalness, resulting in a well-refined personalized TOD dataset. Despite the highly automated process, our dataset demonstrates higher human preference in both user experience (\S~\ref{sec:human_task1})  and personalization (\S~\ref{sec:different_method}) compared to other datasets and methods. From a label alignment perspective, analysis with DST and policy modules show that PicPersona-TOD maintains information accuracy. Specifically, we present an NLG model called \textbf{Pictor}, which demonstrates the ability to generate robustness in personalization, even in unseen domains.

In summary, the contributions of this work are threefold: first, we introduce PicPersona-TOD, a novel TOD dataset that integrates user images into personas and provides personalized system responses. Second, we present a highly automated dataset generation framework that efficiently creates realistic and personalized datasets. Third, through human evaluation, we demonstrate that our dataset enhances user experience through personalization, with benchmark results confirming that personalization does not compromise performance in other critical tasks.

\section{PicPersona-TOD Dataset}

In this section, we introduce \textbf{PicPersona-TOD}, the first personalized TOD dataset based on user image persona. To construct a high-quality personalized TOD dataset, we hypothesize that it should meet three criteria: (1) the user's utterances should be consistent with their image, (2) the system's responses should be appropriately personalized to the user image, and (3) the synthesized dataset should align with the sub-task labels of TOD tasks, such as DST and dialogue policy prediction, while maintaining the information. To address these criteria, our dataset construction pipeline comprises five key stages:  (1) user image collection and dialogue dataset extension, (2) user image and utterance alignment, (3) user utterance style transfer, (4) system response personalization, and (5) data filtering.  For data construction, we mainly employed GPT-4o \cite{gpt4} as the primary language model. The overall process is illustrated in Figure ~\ref{fig:main} and the used prompts are  in Appendix~\ref{sec:appendix_prompt}.

% , thereby assessing its quality as a personalized dialogue dataset and information consistency.

\subsection{Collecting Images and Extending Dialogue}
\label{sec:method-collect}Initially, we select suitable user images that convey sufficient persona information. In order to effectively represent a user persona, each image should be a single person who is positioned in the center of the image, and close enough so that facial detail and clothing information are clearly visible. Based on these criteria,  we selected the Flickr-Faces-HQ \cite{FFHQA} as an image source, and made sure to exclude toddlers, as they are too young to engage in TOD interactions. After collecting the data, we used LLM to extract additional metadata for each image, including estimated age, gender, and formality.

For the dialogue dataset, we combined the MultiWOZ-2.2 \cite{mwoz22} and SGD \cite{sgd} datasets, which include 8,438 and 11,398 dialogues containing 18 service domains in total\footnote{For the generalization test, we excluded the bus, home, and movie domains.}. Additionally, since the movies, restaurants, hotels, and attractions in the dataset exist in the real world, we extended the dataset by collecting Google Maps reviews and Wikipedia entries for each location. Specifically, we added 2,474 sentences from 342 Wikipedia entries and 3,483 reviews from 406 locations on Google Maps. These results were added to the database to reduce hallucination in personalization.

\subsection{Image and Dialogue Data Alignment}
\label{sec:method-align}

After selecting the image and dialogue datasets, we conducted user image and utterance alignment. Since the dialogue datasets lack details like age or gender, we chose emotion as a common attribute, as it is consistently present in both images and dialogues.  We used a fine-tuned emotion classification model\footnote{Fine-tuned model from Hugging Face, \url{cardiffnlp/twitter-roberta-base-sentiment-latest}} to classify emotions in dialogues and prompted an LLM to classify images (positive, neutral, negative), and created image-dialogue pairs if they had the same predicted emotion label. The distribution of emotions across the dataset was 50.92\% for positive, 52.44\% for neutral, and 0.55\% for negative.

\subsection{Alignment User Utterance Style to Image}
\label{sec:method-user}

Next, we adapted the user utterances style to more closely align with the corresponding user images. We performed style transfer by prompting with user images, considering factors such as age, gender, emotion, and contextual cues. Formally, for each $i-th$ dialogue $D_i = (u_0, s_0, u_1, s_1 ... u_T, s_T)$, and its associated image $\text{Img}_i$, we generate a revised utterance $\tilde{u}_t$ at turn $t$; $\tilde{u}_t = \text{LLM} (s_{t-1}, u_{t}, \text{Img}_i)$ where $T$ is the total number of turns, $u$ is user utterance and $s$ is system utterance in the dialogue.

% we modify the tone of the user’s utterances to ensure consistency with their associated image by adjusting utterance styles. This is achieved through

%

\subsection{System Response Style Personalization}
\label{sec:method-system}

In personalizing the system’s responses, we categorized the process into three types and guided the prompts using first impressions and dialogue policy. `Basic Personalization' was applied in most cases, while `Greeting Personalization' was used for dialogue actions involving greetings. For recommendation-related actions, we implemented `Recommendation Personalization', to provide less hallucinated suggestions.

\noindent
\textbf{Basic Personalization} In order to generate personalized system utterances that align with the user's image, we use the \textit{first impression} as a guide for the prompt. This process draws inspiration from human cognitive mechanisms in communication, which consist of two key steps.  First, humans unconsciously infer a first impression of others within milliseconds \cite{psy3_50, psy1_100}, and then adjust their communication tone and style to align with this impression, using it as an inferred persona \cite{psy_book}. Similar to this process, we first generate an impression from the user’s image ($\text{Img}_i$) and then generate personalized system’s utterances in terms of formality, age sensitivity, and emotional context, based on the inferred persona. Specifically, for a given dialogue $D_i$, $\text{Img}_i$ and inferred $\text{Imp}_i$, the personalized system utterance $\tilde{s}_t$ is generated as follows; $\tilde{s}_t = \text{LLM} (s_{t},\tilde{u}_{t}, \text{Img}_i,\text{Imp}_i).$

\noindent
\textbf{Greeting Personalization} Since greetings and closing remarks play a crucial role in creating personal interactions between speakers \cite{mclean2021alexa, greetings}, we specifically tailor the system's greetings and farewells to provide a more engaging and personalized experience. This is achieved by prompting LLM to incorporate specific comments about the user’s appearance, such as mentioning a distinctive feature of their outfit (e.g., ``Nice red hat!" or         ``Congratulations on your achievement") (Figure~\ref{fig:main}, Top-Right).

\noindent
\textbf{Recommendation Personalization} TOD systems often make recommendations, such as \texttt{``How about [location]?"}. In our preliminary experiments, we observed that when the model attempted to personalize these recommendations, it sometimes introduced hallucinated information (e.g., \texttt{``The [location] currently has a festival you might like''}). To mitigate these hallucinations, we performed retrieval-augmented generation \cite{rag} by enhancing the prompt with authentic information gathered from online sources. Specifically, from the reviews about \texttt{[location]} in the database (DB) (\S~\ref{sec:method-collect}), we retrieve the three reviews with the highest cosine similarity to $\tilde{u_t}$, by embedding them using Sentence-BERT\cite{sentencebert}. These were incorporated into the prompt to guide the model in generating factually grounded responses(Figure~\ref{fig:main}, Bottom-Right). After these process we get the personalized TOD dialogue $\tilde{D_i}$ = $\{\tilde{s}_{0:T}, \tilde{u}_{0:T}, \text{Img}_i, \text{Imp}_i\}$.

\subsection{Dataset Quality Control by Filtering}
\label{sec:method-filtering}
\begin{figure}[ht]
    \centering
    \includegraphics[width=0.9\linewidth]{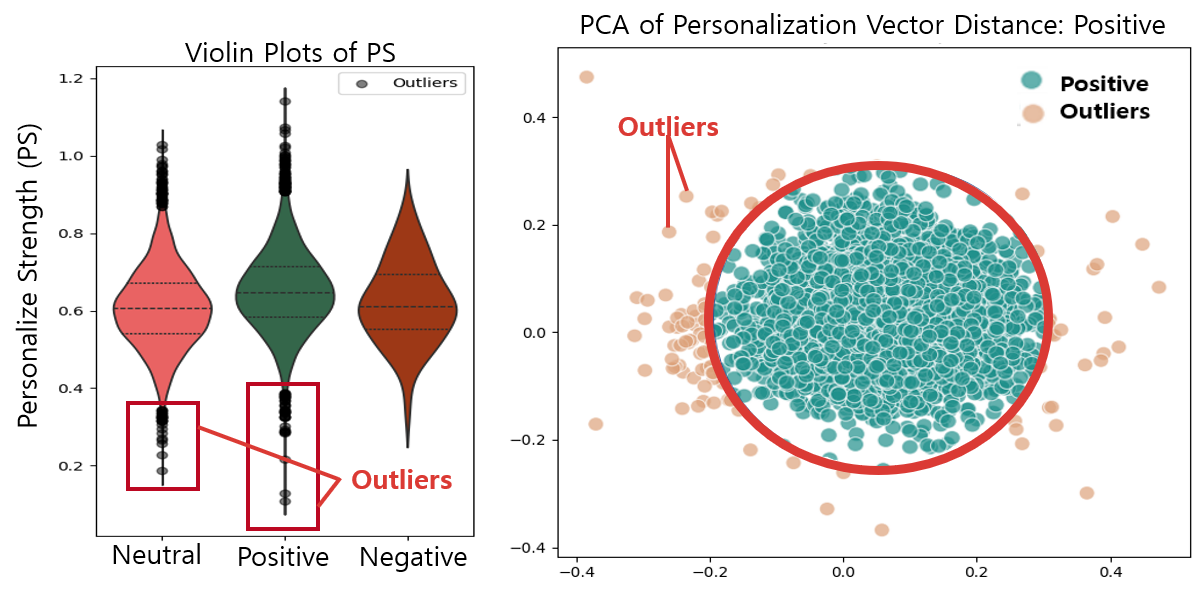}
    \caption{Visualizations of personalization strength and personalization direction filtering processes are shown on the left and right, respectively.}
    \label{fig:vis_filtering}
\end{figure}

Despite the meticulous construction process in the previous subsections, some dialogues may still generate inappropriate utterance style, contain semantic inconsistencies, or lack overall naturalness. To address these potential issues and enhance the overall quality of the dataset, we implement several filtering processes.

\noindent
\textbf{Style Strength Filtering} Dialogues are eliminated if the degree of personalization is too low (Figure~\ref{fig:vis_filtering}, left). To do this, we calculate the strength of personalization for each dialogue $i$ by defining the Personalization Strength ($\text{PS}_i$) as the average distance between the original system response $E(s_t)$ and the personalized response $E(\tilde{s}_t)$: $\text{PS}_i = \frac{1}{T} \sum_{t=0}^{T} \text{Dist}\left( E(\tilde{s}_t), E(s_t)\right)$, where Dist represents the Euclidean distance, $E$ represents embedding with Sentence-BERT. Next, we collect the $\text{PS}$ values for each metadata class (e.g., young, senior) and remove dialogues with $\text{PS}$ values below the threshold, defined as less than $2.5 \times \text{IQR}$ (interquartile range). As a result, 1.49\% of the dataset was filtered out. 

\noindent
\textbf{Style Direction Filtering}
We remove outliers that have a different direction of personalization within the same metadata class (Figure~\ref{fig:vis_filtering}, right). We calculate the Personalization Vector (PV) for each dialogue $i$ as $\text{PV}_i = \frac{1}{T} \sum_{t=0}^{T} \left(E(\tilde{s}_t) - E(s_t)\right)$.  Then, we compute the mean personalization vector for each metadata class, $\text{PV}_{class}$, by averaging the PV vectors within that class. To detect outliers, we calculate the distance ($\text{PD}_i$) between the class mean and the personalization vector of each dialogue : $\text{PD}_i$ = $\text{Dist}(\text{PV}_{class}, \text{PV}_i)$. We define outliers as those with an exceptionally large distance from the mean of the class style vector. We set the threshold as $4.5 \times \text{IQR}$, resulting in the removal of 1.98\% of the dataset.

\noindent
\textbf{Semantic Filtering} 
We filter out semantically misaligned user and system utterances by comparing them with the corresponding DST and dialogue policy labels. For user utterances, we check their alignment with the DST labels. For example, if the label is \texttt{hotel-east, restaurant-expensive}, the user’s utterance should reflect this, such as by saying, ``I need a hotel in the east and an expensive restaurant." Similarly, we verify that system responses align with the dialogue policy label. Semantically misaligned data were filtered by prompting the LLM to check for inconsistencies, resulting in the removal of 2.37\% of the dataset.

% For instance, if the label is \texttt{Inform-restaurant-nandos}, the system’s response should be, `There is an expensive restaurant called Nandos.' 

\noindent
\textbf{Overall Naturalness Filtering}
Lastly, we filter out dialogues that do not exhibit naturalness. Since the system and user utterances are generated turn by turn, some parts of the dialogue may not flow naturally. To remove such unnatural instances, we provide the entire dialogue to the LLM to assess its flow. As a result, we remove 4.39\% of the dialogues. After these individual filtering stages, 92.59\% of the initial dataset is retained.

% We provide case studies on personalization and filtered results in Appendix~\ref{sec:appendix_case}, along with demographic details in Appendix~\ref{sec:appendix_demo}.

\begin{figure*}[t!]
    \centering
    \includegraphics[width=\linewidth]{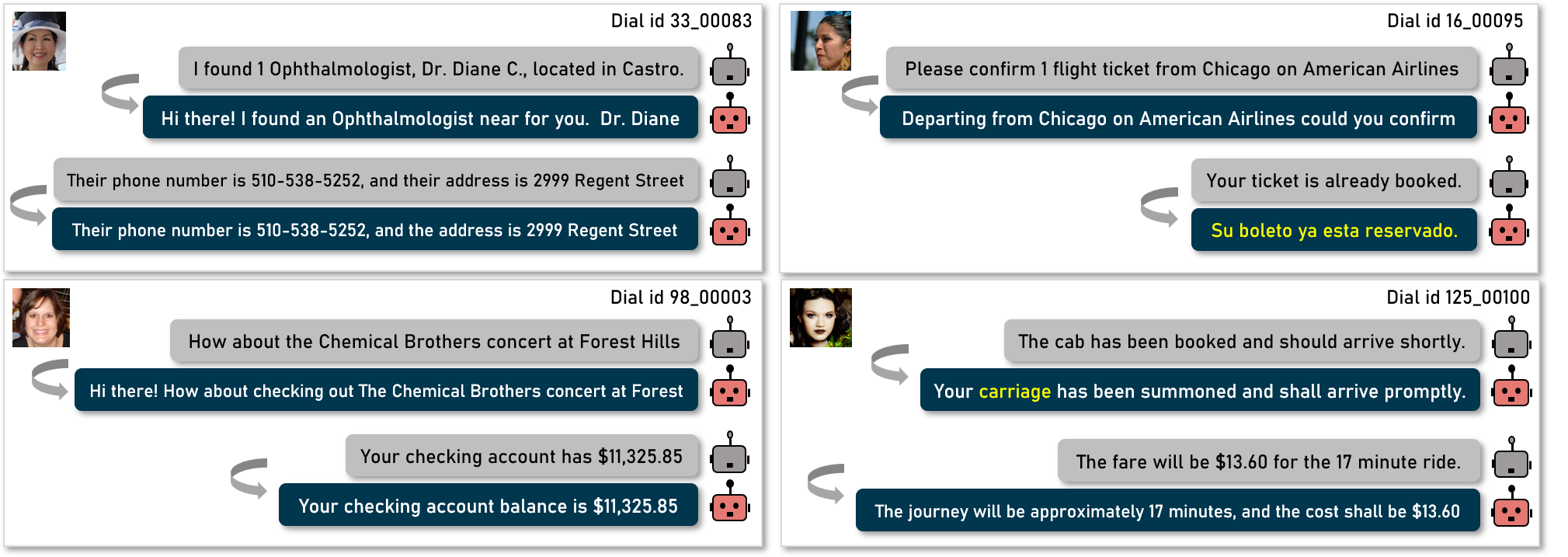}
    \caption{Examples of filtered-out results: style strength filtering (left) and style direction filtering (right).}
    \label{fig:case2}
\end{figure*}
\subsection{Case Study for Filtering}
Figure~\ref{fig:case2} shows examples of filtered-out results for the style strength filter (left) and the style direction filter (right). The style strength filter removes instances with minimal or unchanged personalization, while the style direction filter excludes cases where personalization led to inappropriate changes, such as medieval-style language or switching to a different language.

\section{PicPersona-TOD Analysis}
\begin{table*}[t]
\centering
\resizebox{\textwidth}{!}{%
\begin{tabular}{lcccccccccc}\hline
 Dataset & Persona Mod.      & Dialogue Mod. & Dialogue Type            & Subtask                & Collection                           & \# of Dial & \# of Serv. & Avg Turn & Avg Tok \\\hline
MultiWoZ \cite{mwoz20}            & -            & Text              & TOD           & DST, Pol                & Human                              & 8,438   & 7             & 13.46 & 13.13       \\
ABCD \cite{ABCD}                & -            & Text              & TOD           & Pol                     & Human                               & 8,034   & 30            & 22.08 & 9.17        \\
SGD \cite{sgd}                 & -            & Text              & TOD           & DST, Pol                & Bot+Human                            & 16,142  & 16            & 20.44 & 9.75        \\
STAR \cite{mosig2020star}              & -            & Text              & TOD           & Pol                     & Human                              & 5,820   & 13            & 21.71 & 11.2        \\
TOAD  \cite{toad}               & Text         & Text              & TOD           & DST, Pol                & GPT3.5                              & 8,087   & 11            & 9.23  & 10.6        \\
SIMMC-2.0 \cite{kottur2021simmc} & -            & Text, Vision      & TOD           & Disamb, Coref., DST     & Bot+Human                            & 11,244  & 2             & 10.4  & 13.7        \\
DialogCC  \cite{lee2024dialogcc}  & -            & Text, Vision      & Open    & Image Ret, Response Pred. & GPT4, CLIP                           & 83k     & -             & 8.20   & -           \\
MPChat \cite{mpchat}              & Text, Vision & Text, Vision      & Open    & Image Ret.             & Reddit                            & 15k     & -             & 2.85     & 18.5        \\
STARK \cite{lee2024stark}
& Text, Vision & Text, Vision      & Open    & Image Ret.             & GPT4, Diffusion & 0.5M        & -               & 5.30   & -           \\
\rowcolor[gray]{0.9} PicPersona-TOD \textbf{(ours)}           & \textbf{Text, Vision} & Text, Vision      & TOD           & DST, Pol                & GPT4,Google Map, Wiki     & \textbf{18,148 }   & \textbf{18}             & 17.23     & 12.67       \\\hline
\end{tabular}%
}
\caption{Comparison of statistics with other datasets in terms of personalization and modality. 'Mod.', 'Serv.', and 'Pol.' stand for modality, service, and policy prediction, respectively.}
\label{tab:dataset_compare}
\end{table*}

In this section, we analyze our dataset in comparison to other datasets(\S~\ref{sec:previous_dataset}), followed by an analysis across key dimensions such as word difficulty, politeness (\S~\ref{sec:lexical}), and emotions (\S~\ref{sec:emotion}).

\subsection{Comparison with Existing Datasets}
\label{sec:previous_dataset}

% In Table~\ref{tab:dataset_compare},  we compare PicPersona-TOD with other datasets in terms of personalization and dialogue data modality. As can be seen in the results, our dataset holds a unique position compared to other datasets, as it is the only TOD dataset that incorporates a visionary persona. Additionally, we integrate external knowledge from online platforms, enabling richer and less hallucinated personalization, which is a feature not found in other personalized datasets. Furthermore, our dataset contains a considerable amount of dialogue data and covers a wide range of services compared to the other TOD datasets.

% In Table~\ref{tab:dataset_compare}, we compare PicPersona-TOD with other datasets in terms of personalization and dialogue data modality. As can be seen in the results, our dataset holds a unique position as the only TOD dataset that incorporates a visionary persona. Additionally, by integrating multiple TOD datasets, it covers a wide range of services with a substantial amount of dialogue data. We also enriched the personalization by external sources such as Google Reviews and Wiki information.

In Table~\ref{tab:dataset_compare}, we compare PicPersona-TOD with other datasets in terms of personalization and dialogue data modality. As shown in the results, our dataset holds a unique position as the only TOD dataset that incorporates a visionary persona. Additionally, by integrating multiple TOD datasets, it covers a wide range of services with a substantial amount of dialogue data. We also enhance personalization through the incorporation of external sources, such as Google Reviews and Wiki information, which constitutes a notable aspect of our dataset.

\subsection{Word Complexity \& Politeness}
\label{sec:lexical}
\begin{figure}[]
    \centering
    \includegraphics[width=0.95\linewidth]{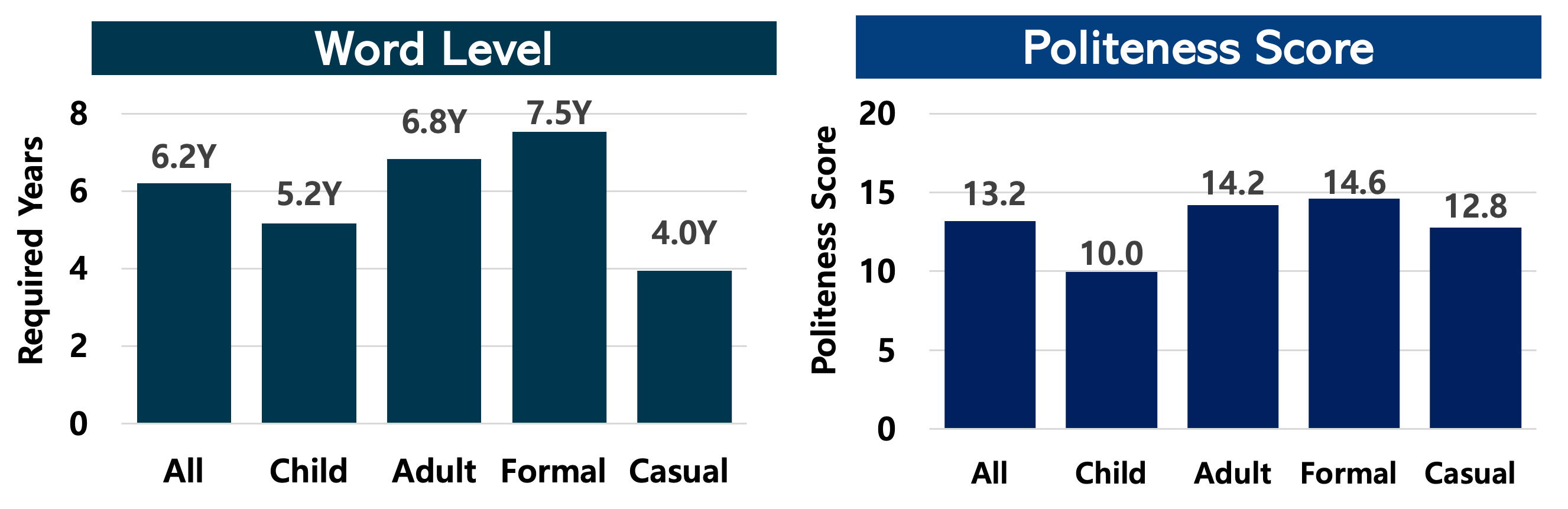}
    \caption{Lexical analysis of PicPersona-TOD. "Word level" refers to the required years of education needed to understand, while the politeness score represents the average use of politeness strategies.}
    \label{fig:lexical}
\end{figure}

\noindent
\textbf{Word Complexity} We analyzed the complexity of system responses across different user scenarios, including interactions with children, adults, and within formal and informal contexts (Figure \ref{fig:lexical}). To assess lexical difficulty, we evaluate the number of years of education required to comprehend the system utterance\footnote{We used the Gunning Fog Scale from the Textstat Python library \url{https://pypi.org/project/textstat/}. }. On average, the system's responses require 6.2 years of education to be understood. Responses for children require 5.2 years of education, while those for adults require 6.17 years. This indicates that the system effectively personalized its word choice based on the user's age. 
% Additionally, the system tailors its language based on the formality of the context; casual responses require 4.0 years of education, while formal responses demand 7.5 years. This adaptive variation highlights the system's effective personalization as it adjusts its language complexity according to the user's age and the contextual formality.

\noindent
 \textbf{Politeness} Politeness of responses is assessed by measuring the average number of politeness strategies used per sentence, where higher scores indicate greater politeness\footnote{We used the PolitenessStrategies library from ConvoKit \url{https://convokit.cornell.edu/}, and measure the average number of strategy.}. Overall, PicPersona-TOD achieved a politeness score of 13.2. In formal contexts, the score rises to 14.6, while in casual contexts it drops to 12.8. For children, the politeness score decreases to 10.0, while for adults, it increases to 14.2. This demonstrates the system's ability to adjust its politeness based on the user’s context.
 
 % These results show that PicPersona-TOD provides personalized responses with appropriate word complexity and politeness based on the user.

\subsection{Emotion Awareness in Responses}
\label{sec:emotion}

Table \ref{tab:emotion} compares the distribution of emotions identified in the system responses across the MultiWoZ and PicPersona-TOD datasets, categorized into 27 emotions using GoEmotions' taxonomy \cite{goemotion} with GPT-4 as the classifier. While the neutral emotion is most common in both datasets, PicPersona-TOD exhibits a significantly lower proportion of neutral responses (61.50\%) compared to MultiWoZ (74.97\%), which indicates a more emotive reaction. Furthermore, we analyze how the system personalizes its responses based on the user's emotional state (third column of Table \ref{tab:emotion}). We found that when the user's image has positive emotions, the system responds with a broader range of emotions, such as joy and gratitude. In contrast, when the user's emotions are neutral or negative, the system tends to generate more neutral responses, with a greater emphasis on empathy and care.

\begin{table}[h]
\centering
\resizebox{\columnwidth}{!}{%

\begin{tabular}{lr|lr|lr}
\hline
\multicolumn{2}{c|}{MultiWoZ} & \multicolumn{2}{c|}{PicPersona-TOD} & \multicolumn{2}{c}{PicPersona-TOD (pos)}     \\ \hline
neutral           & 74.97     & neutral          & 61.50     & neutral              & 44.66              \\
curiosity         & 9.18      & curiosity        & 11.79     & approval              & 16.51               \\
gratitude         & 6.31      & approval         & 8.65      & curiosity              & 11.65              \\
approval          & 3.56      & gratitude        & 6.52      & gratitude             & 7.77              \\
optimism          & 1.26      & annoyance        & 2.26      & excitement             & 4.86               \\
apology           & 0.92      & optimism         & 1.88      & admiration                    &  2.43            \\ \cline{5-6} 
annoyance         & 0.69      & excitement       & 1.88      & \multicolumn{2}{l}{PicPersona-TOD (neu/neg)} \\ \cline{5-6} 
confusion         & 0.57      & admiration       & 1.00      & neutral              & 66.69              \\
caring            & 0.57      & amusement        & 0.63      & curiosity              & 11.84               \\
disappointed      & 0.57      & disappointed     & 0.63      & gratitude              & 6.10                \\
joy               & 0.35      & caring           & 0.51      & approval                &  5.90               \\
excitement        & 0.35      & joy              & 0.38      & caring                & 2.54               \\
admiration        & 0.23      & confusion        & 0.38      & optimism              & 1.36\\ \hline
\end{tabular}

}
\caption{The proportion (\%) of the most frequent emotions in system responses within the test dialogue.}
\label{tab:emotion}
\end{table}

% \caption{The ratio  of Top-k emotions of system response in the test dialogue.}

% 예진쓰를 반영하여 고쳤으나 다시 읽어보기

\section{Human Evaluation}
\subsection{Evaluation for Quality}
\begin{figure}[h]
    \centering
    \includegraphics[width=0.95\linewidth]{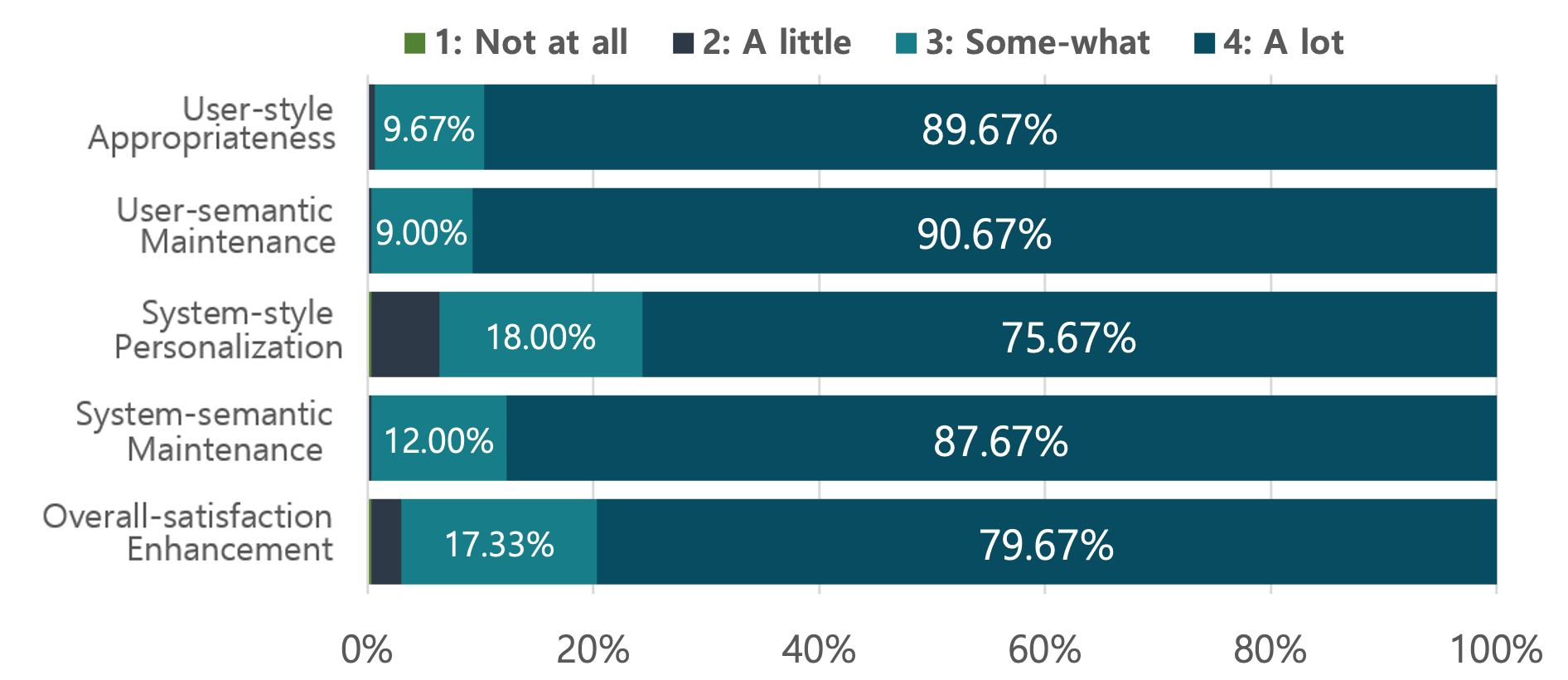}
    \caption{Human evaluation results for dataset quality using a 4-point Likert scale.}
    \label{fig:human_eval_1}
\end{figure}
\label{sec:human_task1}
To assess user satisfaction with personalized results, we conducted human evaluation that focused on verifying the degree of personalized style and information retention. Three evaluators rated 100 randomly selected dialogues on a 4-point Likert scale and measured the scores for both user and system utterances using five questions (Appendix~\ref{sec:appendix:humaneval_detail}). The results were highly positive, with average scores of 3.89 for user style appropriateness, 3.90 for user semantic consistency, 3.69 for system style personalization, 3.87 for system semantic consistency, and 3.76 for overall user satisfaction (Figure~\ref{fig:human_eval_1}). Additionally, we observed a strong inter-rater agreement of 0.85, measured using Krippendorff's Alpha. These results confirm that the PicPersona-TOD meets our predefined criteria: (1) user and image consistency, (2) personalized system responses, and (3) maintaining the original information. We also performed the same evaluation using GPT-4, which showed a high inter-rater correlation with human evaluators, achieving a score of 0.84 (Appendix~\ref{sec:appendix:humaneval_detail1}).

\subsection{Other Personalization Methods}
\label{sec:different_method}
We conducted a comparative evaluation of PicPersona-TOD against two other personalized methods. Since no method currently exists for incorporating visual persona into TOD, we compared PicPersona-TOD with methods that rely on textual modalities. The first baseline, \citet{toad}, personalizes dialogue by mirroring the user's noun and verb phrases, while the second method, \citet{joshi2017personalization}, personalizes interactions based on age and gender information. We sampled 120 dialogues from various scenarios and had three human judges evaluate them to determine the superior method based on personalization quality (Appendix~\ref{sec:appendix:humaneval_detail2}). As shown in Table \ref{tab:different_method}, PicPersona-TOD consistently outperformed the text-based methods across diverse user scenarios.  This result emphasizes the importance of rich, concurrent image personas for personalization, compared to relying on textual personas.

\begin{table}[]
\centering
\resizebox{0.95\columnwidth}{!}{%
\begin{tabular}{l crrrrrr}
\hline
 & \multirow{2}{*}{All} & \multicolumn{3}{c}{Age} & &\multicolumn{2}{c}{Emotion}  \\
\cline{3-5}\cline{7-8}
 && Senior & Adult & Child & & Pos & Neu\&Neg  \\
\hline
 \makecell[l]{\citet{toad}}& 2.22 & 0.00 & 1.80 & 4.17 && 1.01 & 3.70  \\
Tie & 13.33 & 9.52 & 14.41 & 12.50 && 6.06 & 22.22 \\
PicPersona-TOD & \textbf{84.44} & \textbf{90.48} & \textbf{83.78} & \textbf{83.33} && \textbf{92.93} & \textbf{74.07} \\
\hline
 \makecell[l]{\citet{joshi2017personalization}}
& 9.44 &  14.29 & 9.01 & 8.33 && 4.04 & 16.05 \\
Tie & 23.33 & 33.33 & 22.52 & 20.83 && 15.15 & 33.33 \\
PicPersona-TOD & \textbf{67.22} & \textbf{52.38} & \textbf{68.47} & \textbf{70.83} && \textbf{80.81} & \textbf{50.62} \\
\hline
\end{tabular}
}
\caption{The winning ratio (\%) when comparing PicPersona-TOD with personalization methods. }
\label{tab:different_method}
\end{table}

\begin{figure}[h]
    \centering
    \includegraphics[width=\linewidth]{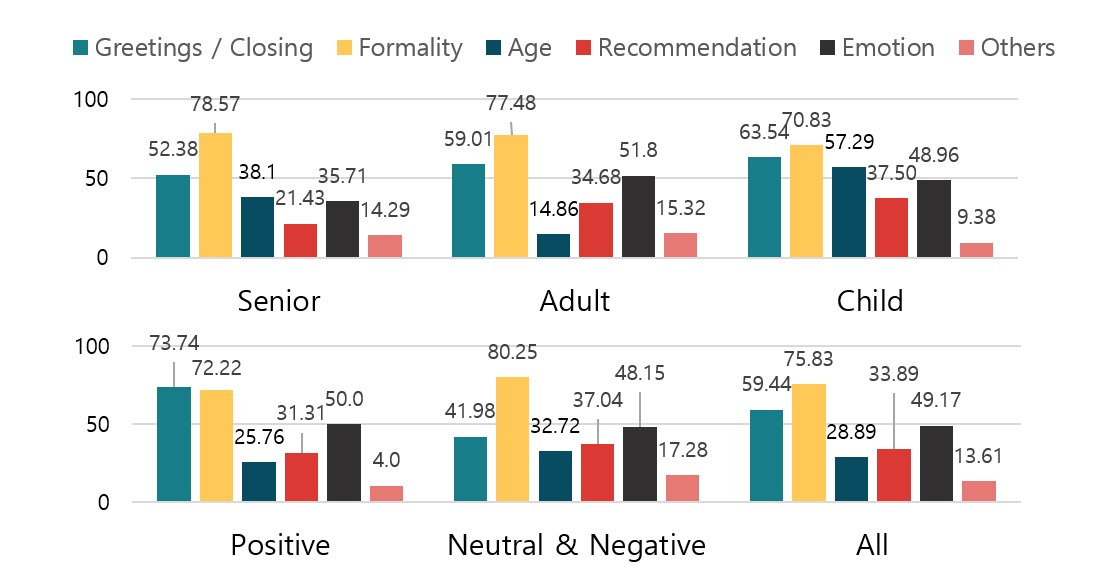}
    \caption{The winning characteristics analysis across ages and emotions.}
    \label{fig:reasons}
\end{figure}
Figure~\ref{fig:reasons} includes the distribution of key factors that influence the evaluators' preferences related to Table~\ref{tab:different_method}. In all user scenarios, appropriate formality is the most noticeable aspect of personalization for human evaluators. For children and in positive contexts, tailored greetings enhance the sense of personalization, while in neutral and negative environments, emotional awareness serves as a key factor in creating a personalized experience. 

\section{Baselines}
In this section, we introduce \textbf{Pictor}, an NLG baseline trained on the PicPersona-TOD dataset to generate personalized responses. Additionally, we provide models for DST and policy prediction, allowing for comparisons with other datasets.
\subsection{Baseline for NLG}
\begin{figure}
    \centering
    \includegraphics[width=0.95\linewidth]{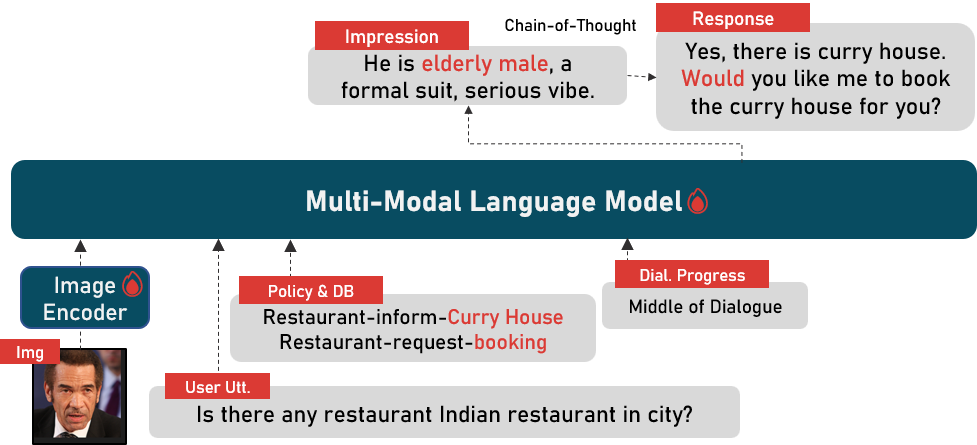}
    \caption{Overview of the proposed model Pictor.}
    \label{fig:baseline_nlg}
\end{figure}
Using the PicPersona-TOD dataset, we developed a multimodal TOD response generation model named \textbf{Pictor} (Figure~\ref{fig:baseline_nlg}). Pictor generates personalized responses ($\tilde{s}_i$) by leveraging user images and dialogue context. Specifically, the input to the Pictor model for turn $t$ in $D_i$ includes the turn progress (e.g., start, middle, end), user's utterance ($\tilde{u}_t$), dialogue policy ($pol_t$), DB results ($\text{DB}_t$), and the user's image ($\text{Img}_i$). Similar to the process used in constructing the PicPersona-TOD dataset, we first generate the user's impression and then, based on this impression, generate a personalized response $\tilde{s}_i$ . Pictor is based on the LLaVA 7B \cite{llava} and 1.5B \cite{tinyllava} models, which are known for strong performance across various vision-language tasks. We train the Pictor model by utilizing a LoRA \cite{lora} adapter with rank of 16. Detailed information is in Appendix~\ref{sec:appendix_training_details}.

\subsection{Baseline for DST and Policy}

The PicPersona-TOD dataset also supports a range of TOD tasks, including DST and policy prediction. To establish baselines for these tasks, we utilized the PPTOD model \cite{pptod} and trained the DST and policy prediction models using both T5-base and T5-small variants \cite{t5}. For the DST task, the input at turn $t$ is defined as the concatenation of all user and system utterances up to and including turn $t$, which can be expressed as $\text{Input}_{\text{DST}, t} = [\tilde{u}_1, \tilde{s}_1, \tilde{u}_2, \tilde{s}_2, \dots, \tilde{u}_t]$.  The output of the DST is represented as slot-value pairs (e.g., \texttt{hotel-name: Green Hotel}).

For the policy prediction task, the input at turn $t$ is similarly constructed, but with the addition of the predicted dialogue state; $\text{Input}_{\text{POL}, t} = [\tilde{u}_1, \tilde{s}_1, \tilde{u}_2, \tilde{s}_2, \dots, \tilde{u}_t, \text{DST}_t]$. The policy prediction model generates the appropriate system action, such as a request for further information (e.g., \texttt{Request-restaurant-foodtype}).
\section{Baseline Evaluation}

% We evaluate baseline models by comparing Pictor's NLG performance with other vision-language models, testing it on unseen domains, and conducting an ablation study. Lastly, we assess the DST and policy models to demonstrate information accuracy.

\begin{figure}
    \centering
    \includegraphics[width=\columnwidth]{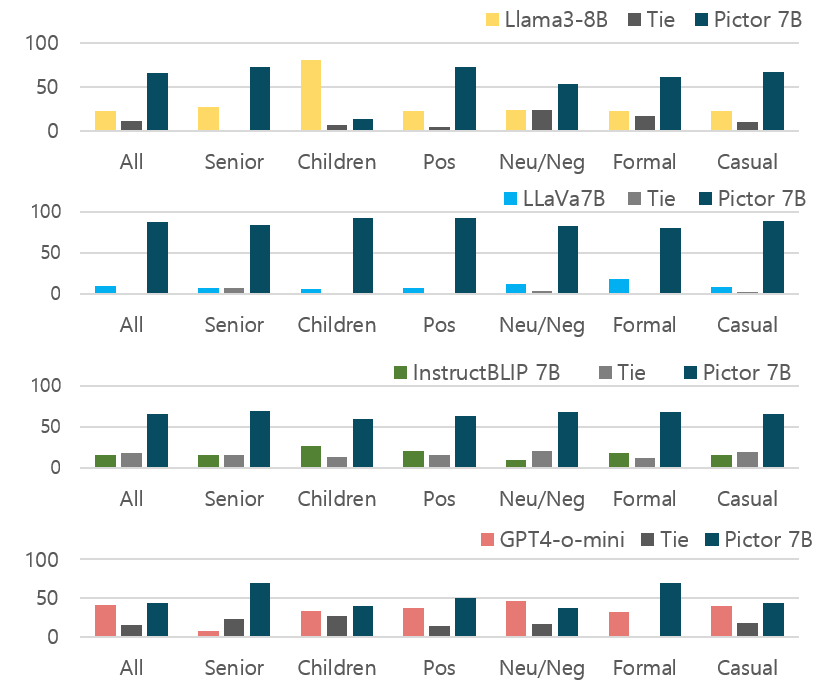}
    \caption{Performance comparison of LLama 3-8B, Pictor 7B with LLaVA 7B, InstrucBLIP 7B, and GPT-4o mini across various user scenarios. }
    \label{fig:LLMs}
\end{figure}
\subsection{Comparison with Other LLMs}
\label{sec:exp_pictor_llm}
We conducted a comprehensive comparison of our Pictor 7B model with several prominent vision-LLMs, including Llama3-8b\cite{llama3} LLaVA 7B, InstructBLIP 7B \cite{instructblip}, and GPT-4o-mini \cite{4omini}. For the evaluation, we sampled 100 dialogues, with assessments performed by the GPT-4 model. As illustrated in Figure \ref{fig:LLMs}, Pictor consistently outperforms the other sLLM models in terms of personalization quality across categories. In comparison to GPT-4o-mini, which likely has more parameters than Pictor, Pictor demonstrates better results, except in neutral/negative cases. These findings highlight the importance of datasets specifically designed for personalization, such as PicPersona-TOD, to achieve optimal performance in personalized scenarios.

\subsection{Generalization Performance Evaluation}
\label{sec:exp_unseen}
\begin{table}[h]
\centering
\resizebox{\columnwidth}{!}{%
\begin{tabular}{lccccc}
\hline
Domain         & Natural. &  Fluency &Personalize &Semantic & User Satisfaction \\\hline
BUS            & 3.67                                                   & 3.79                                               & 3.51                                                 & 3.78                                                & 3.70                         \\
MOVIE          & 3.72                                                   & 3.79                                               & 3.58                                                 & 3.82                                                & 3.81                        \\
HOME           & 3.90                                                    & 3.96                                               & 3.78                                                 & 3.93                                                & 3.88                        \\\hline
\end{tabular}%
}
\caption{Human evaluation for of the Pictor in unseen domains.}

\label{tab:generalization}
\end{table}

In Table~\ref{tab:generalization}, we perform the human evaluation on the generalization performance of the Pictor model across the Bus, Movie, and Home domains from the SGD dataset, which were not included in the model's training data. Note that these datasets were constructed using the same process as the dialogues in the PicPersona dataset, which also includes user images. We sampled 100 dialogues from each domain and performed zero-shot inference, followed by human evaluations conducted by three annotators using a 4-point Likert scale (Appendix~\ref{sec:appendix:humaneval_detail1}). The results demonstrate that, despite these domains not being part of Pictor's training set, the model is able to achieve strong personalization with strong user satisfaction close to 4. We attribute this to the inclusion of two large-scale TOD datasets as a dataset source, which cover a wide range of domains.
% , enabling the model to generalize unseen domains.

\subsection{Ablation Study on Pictor}
\begin{table}[t]
\resizebox{\columnwidth}{!}{

\begin{tabular}{lrrrr}
\hline
Input & \multicolumn{1}{l}{BLEU} & \multicolumn{1}{l}{Style} & \multicolumn{1}{l}{Semantic} & \multicolumn{1}{l}{Overall} \\ \hline
\textbf{LLaVA 1.5B} &&&& \\
Pol +DB  & 8.75  & 2.71  & 2.95 & 2.60 \\
Pol +DB + $\tilde{u}$ & 14.28 & 3.15 & 3.52 & 3.1 \\
Pol +DB + $\tilde{u}$ + Img & \textbf{16.18} & \textbf{3.47} & 3.74 & \textbf{3.41} \\
\rowcolor[gray]{0.9}Pol +DB + $\tilde{u}$ + Img + Imp (\textbf{Pictor})  & 14.96 & \textbf{3.47} & \textbf{3.76} & \textbf{3.41} \\
\hline
\textbf{LLaVA 7B} &&&& \\
Pol +DB  & 15.46 & 3.00 & 3.49 & 2.99 \\
Pol +DB + $\tilde{u}$ & 20.21 & 3.18 & 3.63 & 3.22 \\
Pol +DB + $\tilde{u}$ + Img & \textbf{22.01} & 3.48 & 3.82 & 3.50 \\
\rowcolor[gray]{0.9}Pol +DB + $\tilde{u}$ + Img + Imp (\textbf{Pictor}) & 20.77 & \textbf{3.51} & \textbf{3.89} & \textbf{3.53} \\ \hline
\end{tabular}
}

\caption{Ablation study for training Pictor model. Each experiment was repeated three times, and the results were averaged.}
\label{tab:ablation}
\end{table}

We conducted an ablation study to examine how different components affect the generation performance of Pictor. We evaluated the model using BLEU scores\footnote{nltk.translate.bleu\_score} and GPT-4 assessments\footnote{Note that the GPT4 scores use the same question formats detailed in \S~\ref{sec:human_task1}.} for semantic accuracy, style, and overall satisfaction (Table \ref{tab:ablation}). Results reveal an interesting finding about impressions; while omitting to generate impressions resulted in higher BLEU scores, incorporating impressions significantly improved personalization. This trend was even more pronounced in larger models, with the style score increasing from 3.48 to 3.51 when impressions were included.

\subsection{DST and Policy Inference Results}
\label{sec:pol_dst}
% \usepackage{multirow}
% \usepackage{graphicx}
\begin{table}[h]
\resizebox{\columnwidth}{!}{
\begin{tabular}{lrrrrrrrr} \hline
\multicolumn{1}{c}{\multirow{2}{*}{Dataset}} &
  \multicolumn{6}{c}{DST} &&
  \multicolumn{1}{c}{Policy} \\
  \cline{2-7} \cline{9-9} 
\multicolumn{1}{c}{} &
  \multicolumn{1}{l}{JGA} &
  \multicolumn{1}{l}{Rest.} &
  \multicolumn{1}{l}{Hotel} &
  \multicolumn{1}{l}{Att.} &
  \multicolumn{1}{l}{Train} &
  \multicolumn{1}{l}{Taxi} &&
  \multicolumn{1}{l}{\small{Entity-F1}} \\ \hline
\multicolumn{8}{l}{\textit{T5-small}}                              \\
Mwoz    & 47.17  & 83.0 & 79.6 & 86.4 & 88.7 & 94.5 && 46.18 \\
PicPersona-TOD & 49.18  & 83.8 & 80.4 & 88.3   & 88.9 & 96.3 && 41.26 \\\cdashline{1-9}
Difference    & $\triangle$ 2.01   & $\triangle$ 0.8  & $\triangle$ 0.8    & $\triangle$ 1.9  & $\triangle$ 0.2  & $\triangle$ 1.8    && $\triangle$ 4.92  \\\hline\hline
\multicolumn{8}{l}{\textit{T5-base}}                               \\ 
Mwoz    & 49.81 & 85.9 & 79.5 & 88.1 & 88.3 & 94.7 && 44.19 \\
PicPersona-TOD & 47.55 & 84.8 & 79.5 & 87.7 & 86.5 & 96.0   && 46.57 \\ \cdashline{1-9}

Difference    & $\triangle$ 2.26  & $\triangle$ 1.1  & $\triangle$ 0.0  & $\triangle$ 0.4  & $\triangle$ 1.8  & $\triangle$ 1.3  && $\triangle$ 2.38 \\
\hline
\end{tabular}%
}
\caption{Result of DST and Policy Inference Tasks. Metric details are in Appendix~\ref{sec:appendix_metric}.}
\label{tab:dst_da}
\end{table}
We conducted experiments to evaluate the information accuracy of the PicPersona-TOD dataset by testing DST and policy models using T5-small and T5-base. For comparison, we also present results using the MultiWOZ dataset. Table~\ref{tab:dst_da} shows that our dataset yields comparable performance to MultiWOZ across most metrics, with only minor differences observed. This consistency suggests that despite the added complexity of our dataset by personalized the user and system, PicPersona-TOD maintain information accuracy on par with human-curated datasets. 
\section{Related Works}
\noindent
\textbf{Advancements in TOD Datasets}
Task-oriented dialogue (TOD) systems have long been a focus of research, with early datasets like ATIS \cite{ATIS}, WOZ2.0 \cite{woz}, and DSTC2 \cite{dstc2} limited to single domains. Later, multi-domain datasets such as M2M \cite{m2m}, MultiWOZ \cite{mwoz20}, SGD \cite{sgd}, and ABCD \cite{ABCD} aimed to improve accuracy but often overlooked user satisfaction. Recent work has sought to enhance user experience by integrating chitchat \cite{chitchat1, chitchat2, chitchat3}, providing detailed explanations in system responses \cite{kim2023injecting, qian2021database}, and considering users' emotional states \cite{cause, feng2024infusing}, though few studies address individual personalization. The most relevant work to our work involves personalization efforts, such as incorporating age and gender \cite{joshi2017personalization} or linguistic patterns \cite{toad}. While some approaches include emotional states \cite{lin2023emous, feng2024infusing}, our method introduces a visionary persona that provides richer and more concurrent information, leading to enhanced user satisfaction.

\noindent
\textbf{Integrating Persona into Dialogue}
While personalized dialogue systems have been widely researched to improve user experience, they have traditionally relied on textual information. Methods include constructing personas through narrative sentences \cite{zhang2018personalizing, zhong2020towards}, key-value pair dictionaries \cite{qian2017assigning, zheng2019personalized}, or users' review histories \cite{kim2024pearl}. Recently, multimodal approaches have emerged, incorporating user images to create richer personas \cite{mpchat, lee2024stark, comicbook_agrawal2023multimodal}. Building on these advances, we introduce a novel method that uses user images as the primary basis for personas in TOD datasets, enabling more contextually appropriate and personalized responses.

\noindent
\textbf{Data Generation by Distillation LLM}
Collecting dialogue data is challenging due to privacy concerns, high costs, and the need for multiple participants. To address this, many works have leveraged LLMs for dataset creation. Examples include compiling seed dialogues \cite{mpchat, kim2022prosocialdialog}, constructing social event graphs \cite{soda}, and generating long-term dialogues \cite{LLM-long1}. Others have used LLMs for commonsense-aware dialogues \cite{chae2023dialogue}, prosocial dialogues \cite{kim2022prosocialdialog}, and task-oriented dialogue (TOD) utterances \cite{kulkarni2024synthdst}. LLM-generated datasets are cost-effective, diverse, and often preferred over human-curated datasets \cite{soda, lee2024stark, lee2021dialogue}. Building on this, we use LLMs to generate personalized, privacy-conscious, and diverse user scenarios.

\section{Conclusion}
In this paper, we have introduced PicPersona-TOD, a novel dataset that personalizes system responses based on a user’s visual persona in the TOD domain. Specifically, PicPersona-TOD incorporates personalized responses in terms of greetings, age, politeness, and emotions. Through user satisfaction experiments, we have demonstrated that PicPersona-TOD enhances personalization while retaining the original information. Additionally, we have proposed and analyzed a baseline model, which includes NLG (Pictor), DST, and policy prediction. Our results show that this method improves personalization without compromising performance in other critical tasks. We believe this work advances research on personalized TOD with multimodal user personas, enabling more natural and human-like interactions.

\section*{Limitations}

\textbf{Lack of Direct Benchmark Comparison} Typically, datasets distilled from LLMs, such as those used for open dialogue \cite{soda} or image-sending in chat \cite{lee2024dialogcc}, are directly compared with traditional, human-created test sets to demonstrate the practical advantages of the new dataset. However, in our case, no traditional, human-made dataset exists for TOD that incorporates user personas, as creating a TOD dataset requires significantly more effort and higher labeling costs compared to open dialogue datasets.

Due to this environment, we were unable to perform direct comparisons with standard datasets. Instead, we evaluated our model’s personalization capabilities against other prominent vision-LLM models in Section~\ref{sec:exp_pictor_llm}, which have been trained on large-scale vision-text datasets. We conducted GPT4 evaluations to assess performance in these comparisons, and the results show a strong preference for the model trained on our dataset. While this does not serve as a direct comparison with a traditional TOD benchmark, we believe it offers a valid alternative, as the results highlight the importance of datasets specifically designed for personalization, such as PicPersona-TOD, showing the advantages of our dataset.

\noindent
\textbf{Limitations in the Use of Other LLMs}
In our study, we conducted dataset curation exclusively using GPT-4o. This decision was based on preliminary experiments, where other vision-LLMs failed to generate personalized responses with the same quality as GPT-4o. However, as open-source vision-LLM models continue to improve, they may become viable options for developing high-quality, cost-efficient dataset-generation pipelines.

\noindent
\textbf{Facial Recognition Requirement} Although our dataset shows strong potential for personalization, its full capability can only be realized in systems equipped with facial recognition technology, such as kiosks or robots. Without such equipment, the image-based personalization features of PicPersona-TOD cannot be effectively utilized.

\noindent
\textbf{Naive Application of Retrieved Results} In the construction of our dataset, we integrated reviews and Wikipedia results as a retrieval-based generation method to enhance system responses. While this approach contributed to improving response quality, it lacked sophistication. Future research could focus on developing more advanced image persona-based retrieval methods, enabling deeper personalization and a more sophisticated understanding of the user’s persona, which would ultimately lead to improved response quality.

% It would be optimal to demonstrate the superiority of our dataset through a comparison with a TOD dataset that incorporates personalization based on user images. However, as no such dataset currently exists, direct comparison was not feasible.

\section*{Ethical Considerations}
In constructing the dataset, we use two sources that involve real users and may raise concerns about privacy and consent. For the image data, we utilize the FFHQ dataset. According to the original paper \cite{FFHQA}, this dataset was created by crawling images from the Flickr website, where only images under permissive licenses were collected. Specifically, the license for these images is CC BY-NC-SA 2.0, which allows for free distribution as long as the use is non-commercial. 

For the Google Map review data, we retrieve reviews using the Google Maps API. Consent from Google Maps users for collecting their data is provided through the Google API Terms of Use, which state that users understand their posts will be publicly available and can be accessed via the API. Additionally, to protect user privacy, we did not collect any personally identifiable information (PII) to ensure anonymity. From a consent perspective, we comply with the licensing terms for both sources and take steps to safeguard user privacy. Furthermore, the use of our data for commercial and for-profit purposes is restricted.

\section*{Acknowledgements}
% should rewrite this

This work was supported by the MSIT (Ministry of Science and ICT), Korea, through the IITP (Institute for Information \& Communications Technology Planning \& Evaluation) grants: RS-2019-II191906 (Artificial Intelligence Graduate School Program at POSTECH, 50\%) and RS-2024-00437866 (ITRC Program 50\%).

\bibliography{reference}
\bibliographystyle{acl_natbib}

\appendix
\begin{figure*}[t!]
    \centering
    \includegraphics[width=\linewidth]{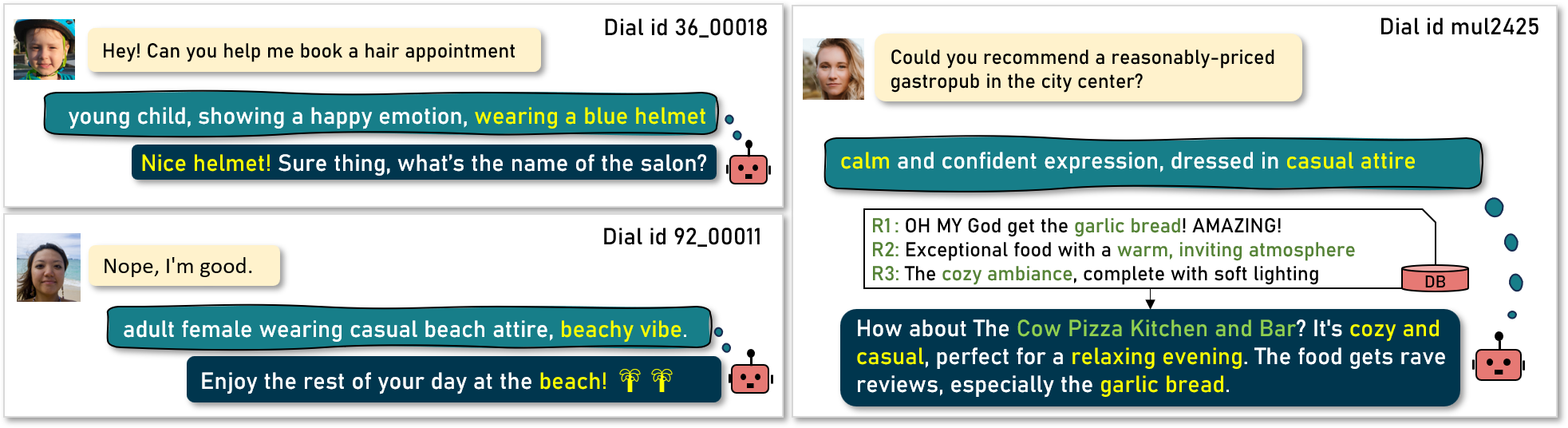}
    \caption{ Examples of greeting personalization (left) and recommendation personalization (right).}
    \label{fig:case1}
\end{figure*}

\label{sec:appendix}

% This can be change 
% \vspace{20\baselineskip}
\newpage

\section{Case Study}
\label{sec:appendix_case}

Figure~\ref{fig:case1} illustrates different methods of system response personalization (\S~\ref{sec:method-system}). For Greetings Personalization (left), the system personalizes greetings and farewells by incorporating specific characteristics from the user's first impressions. For Recommendation Personalization (right), the system retrieves information from reviews and tailors the response based on that information.

% \begin{figure*}[t]
%     \centering
%     \includegraphics[width=\linewidth]{img/casestudy/greeting_DB.png}
%     \caption{ Examples of greeting personalization (left) and recommendation personalization (right).}
%     \label{fig:case1}
% \end{figure*}
% \begin{figure*}[t]
%     \centering
%     \includegraphics[width=\linewidth]{img/casestudy/filtering.png}
%     \caption{Examples of filtered-out results: style strength filtering (left) and style direction filtering (right).}
%     \label{fig:case2}
% \end{figure*}

\section{Demographics}
\label{sec:appendix_demo}
We analyze the distribution of PicPersona-TOD by using V-LLM to obtain information on gender, age, and formality from the user image. The gender distribution shows that 35.82\% of the dataset consists of females, while males represent a slightly larger percentage at 47.71\%.
Regarding age distribution, the adult group constitutes the largest proportion at 74.24\%.
The proportion decreases with age, with users in the child group and the senior group making up 19.66\%, and 6.68\% of the dataset, respectively.
Formality is divided into formal and casual groups, with the casual group accounting for the majority at 86.54\% and the formal group making up 13.46\%. The distribution of emotions across the dataset is 50.92\% for positive, 52.44\% for neutral, and 0.55\% for negative.
These results highlight the diverse demographic representation of the PicPersona-TOD dataset across gender, age, formality, and emotion.

\section{Experiments with GPT-4}
\label{sec:appendix_humantask}
We conducted two parallel evaluation tasks using GPT-4. In the Personalization Quality Evaluation task (\S~\ref{sec:human_task1}), GPT-4 assigned the following scores: Q1: 3.79, Q2: 3.99, Q3: 3.66, Q4: 3.97, and Q5: 3.75, which achieves a high inter-rater reliability with a Krippendorff's alpha of 0.84.

\section{Baseline Training Details}
\label{sec:appendix_training_details}
We trained the Pictor model using both the LLaVA 1.5B and 7B models. For the 1.5B model, we employed LoRA with a rank of 16, an alpha value of 64, a batch size of 16, and a learning rate of 2e-5 over 5 epochs. In the case of the 7B model, LoRA was configured with a rank of 16, an alpha value of 32, a batch size of 16, and a learning rate of 5e-5, also for 3 epochs. Both models employed the Adam optimizer \cite{adam} with no weight decay and utilized a cosine learning rate schedule with a 3\% warmup ratio. All training for the Pictor model was conducted on an NVIDIA A100 GPU.

For the DST and policy models (T5-small and base variants), we used a batch size of 16, learning rate of 1e-3, and trained them for 10 epochs using the AdamW \cite{adamw} optimizer with no weight decay. These models were trained on an NVIDIA A6000 GPU.

\section{Metric for DST and Policy Prediction}
\label{sec:appendix_metric}
When evaluating DST performance, we used two metrics: Joint Goal Accuracy (JGA) and domain-specific JGA. JGA is considered correct if all dialogue states in a turn are accurate. Domain-specific JGA is marked as correct if the dialogue state for the targeted domain is accurate, regardless of other domains \cite{trade}. For Dialogue Policy evaluation, we used Entity F1, which calculates the F1 score for each turn and then averages these scores across all turns \cite{entityF11, entityF12}.

\section{License}
\label{sec:appendix_license}
PicPersona-TOD is synthesized using the MultiWoZ 2.2, SGD, and FFHQ datasets. MultiWoZ 2.2 is released under an MIT license, while SGD is under a CC BY-SA 4.0 license and the images in the FFHQ dataset are licensed under Creative Commons BY 2.0, Creative Commons BY-NC 2.0, Public Domain Mark 1.0, Public Domain CC0 1.0, or U.S. Government Works licenses. These licenses permit free use, copying, modification, and publication for non-commercial purposes. 

\section{Human Evaluation Details}
\label{sec:appendix:humaneval_detail}
For human evaluation, we hired three native English-speaking evaluators through the Upwork\footnote{www.upwork.com} platform. They were informed that all personal information would remain anonymous and that their submitted responses would be used solely for research purposes.

\label{sec:appendix:humaneval_detail}
\subsection{Section~\ref{sec:human_task1} and ~\ref{sec:exp_unseen}}

\label{sec:appendix:humaneval_detail1}
\begin{figure}
    \centering
    \includegraphics[width=0.8\columnwidth]{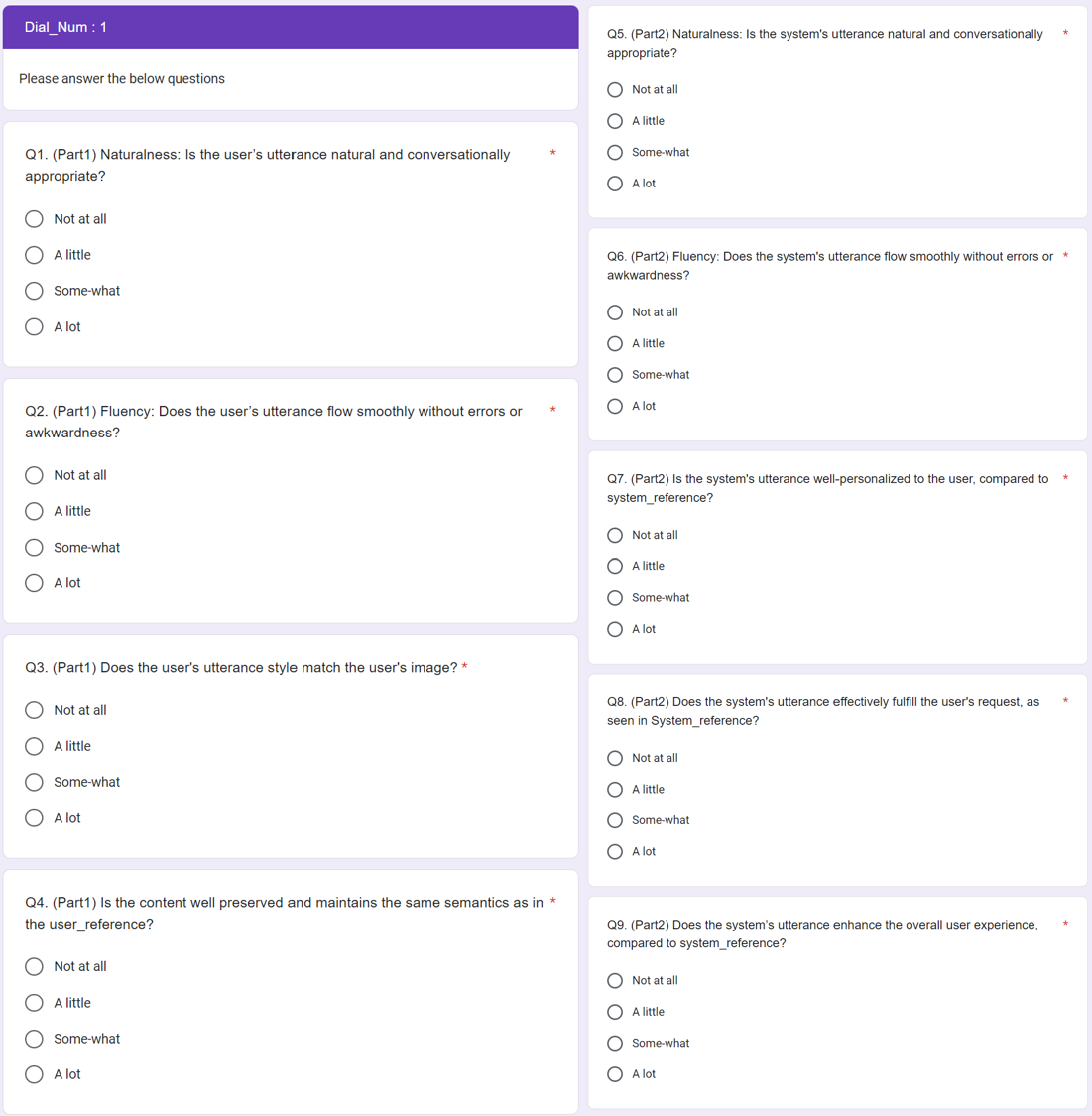}
    \caption{A screenshot of the Personalization Quality Evaluation.}
    \label{fig:screenshot_task1}
\end{figure}

The evaluators were asked to rate the quality of personalization in PicPersona-TOD or Pictor's utterances by selecting one of the provided \textit{Options} in response to the following questions, which pertained to user utterances (U1–U4) and system utterances (S1–S5).

\begin{itemize}
    \small{
    \item \textbf{U1\&S1.} Naturalness: Is the \{user/system\}’s utterance natural and conversationally appropriate?

    \item \textbf{U2\&S2.} Fluency: Does the \{user/system\}’s utterance flow smoothly without errors or awkwardness?
    
    \item \textbf{U3.} Does the user's utterance style match the user's image?

    \item \textbf{U4.} Is the content well preserved and maintains the same semantics as in the original user utterance?

    \item \textbf{S3.} Is the system's utterance well-personalized to the user, compared to the original system utterance?

    \item \textbf{S4.} Does the system's utterance effectively fulfill the user's request, as seen in the original system utterance?

    \item \textbf{S5.} Does the system’s utterance enhance the overall user experience, compared to the original system utterance?

    \item \textbf{Options: } Not at all (1) / A little (2) / Some-what (3) / A lot (4).
    }
\end{itemize}
We used Google Forms for the evaluation, and Figure \ref{fig:screenshot_task1} shows the sample screenshot that the evaluators performed.
The scores corresponding to each option were used to calculate the results.

\subsection{For Section~\ref{sec:different_method}}
\label{sec:appendix:humaneval_detail2}
\begin{figure}
    \centering
    \includegraphics[width=0.8\columnwidth]{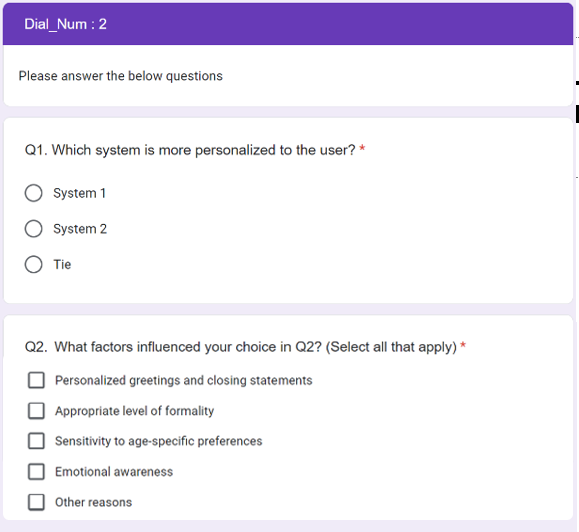}
    \caption{A screenshot of the Comparison with Other Personalization Methods.}
    \label{fig:screenshot_task2}
\end{figure}
The evaluators are asked to compare two anonymous systems, and indicate which part has been improved:

\begin{itemize}
\small{
    \item \textbf{Question:} Which system is more personalized to the user?
    
    \textbf{Options:} System 1 / System 2 / Tie

    \item \textbf{Question:} Is the content well preserved and maintains the same semantics as in the original user utterance?
    
    \textbf{Options:}
    \begin{itemize}
        \item \textbf{Personalized greetings and ending statements:} A personalized response could include customized greetings or closing remarks.

        \item \textbf{Formality:} A personalized response should be appropriately formal or informal, depending on the situation.

        \item \textbf{Age sensitivity:} A personalized response should be age-sensitive.

        \item \textbf{Emotional awareness:} A personalized response should be emotionally aware.

        \item \textbf{Other reasons}
    \end{itemize}
    }
\end{itemize}

Figure \ref{fig:screenshot_task2} illustrates a screenshot of the questions used for evaluation.

\section{Sample of PicPersona-TOD}
\begin{figure*}[h]
    \centering
    \includegraphics[width=\textwidth]{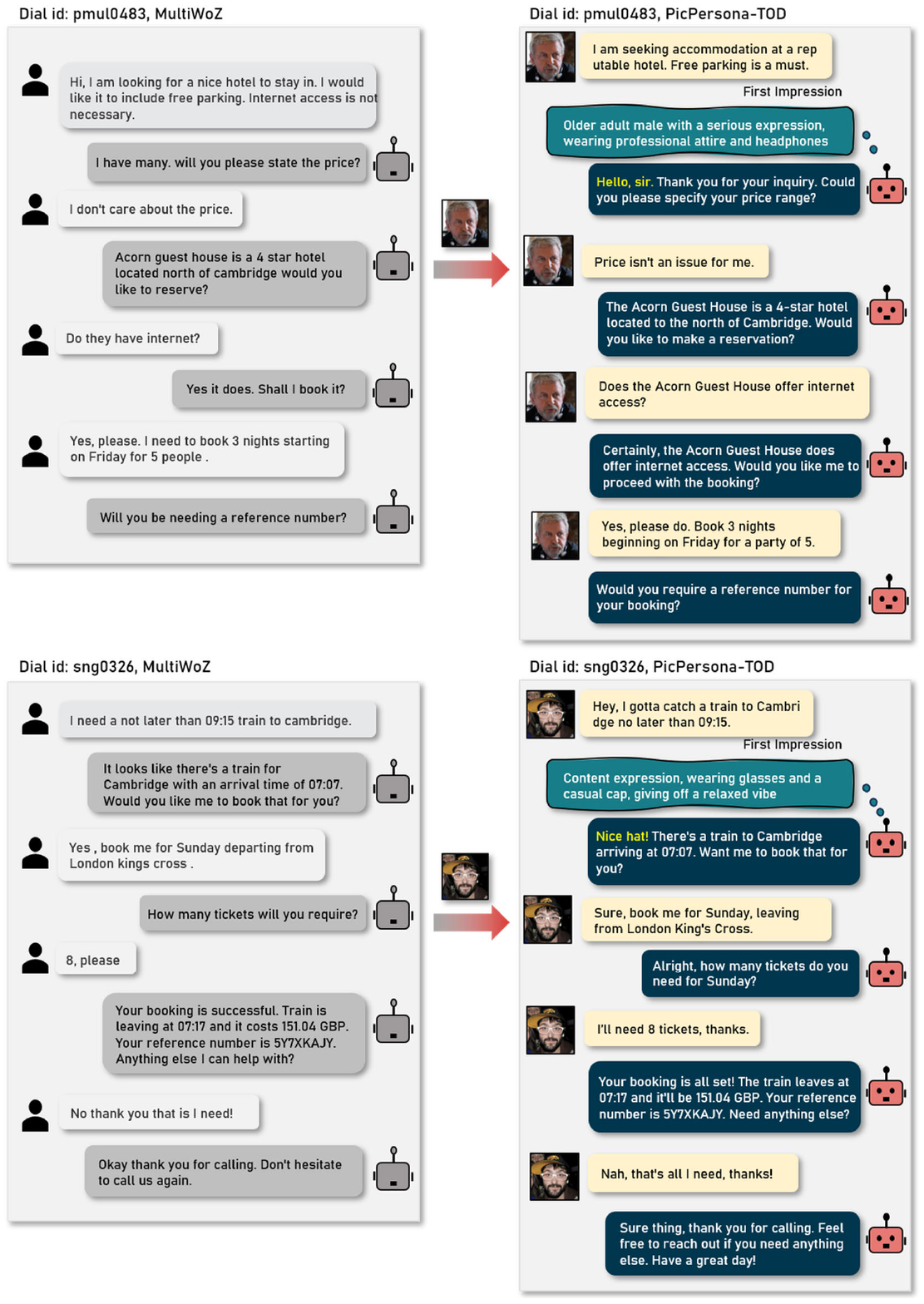}
\caption{PicPersona-TOD dialogue sample.}
\label{fig:sample}
\end{figure*}

Figure \ref{fig:sample} displays samples from the PicPersona-TOD dataset. The left side shows examples from the original MultiWoZ dataset, while the right side presents samples from PicPersona-TOD.

\begin{figure*}[h]
    \centering
    \includegraphics[width=\textwidth]{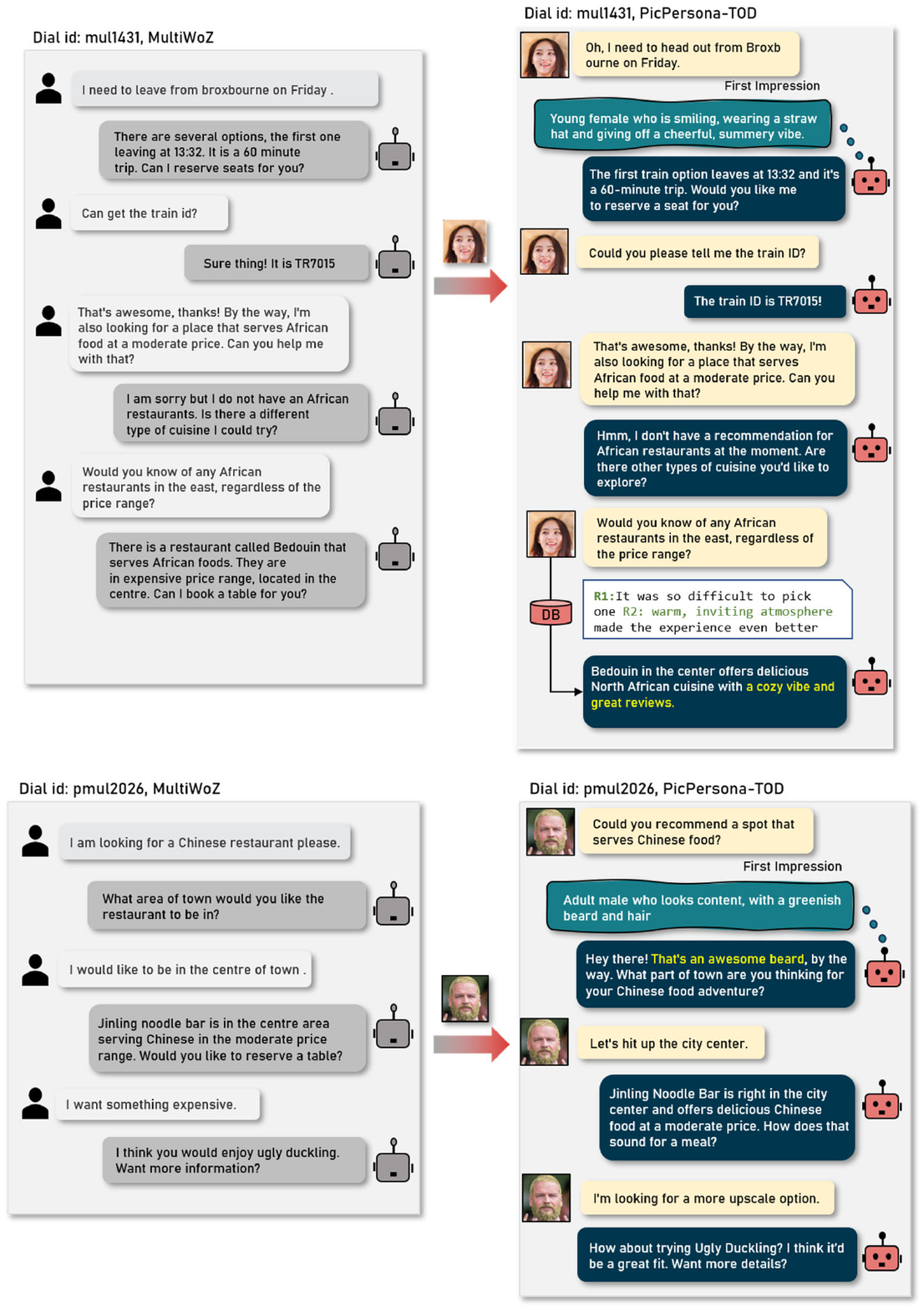}
\end{figure*}

\section{Prompt Templates}
\label{sec:appendix_prompt}
Example prompt templates have been included starting from page 18.

\begin{figure*}[ht]
    \centering
    \includegraphics[width=\textwidth]{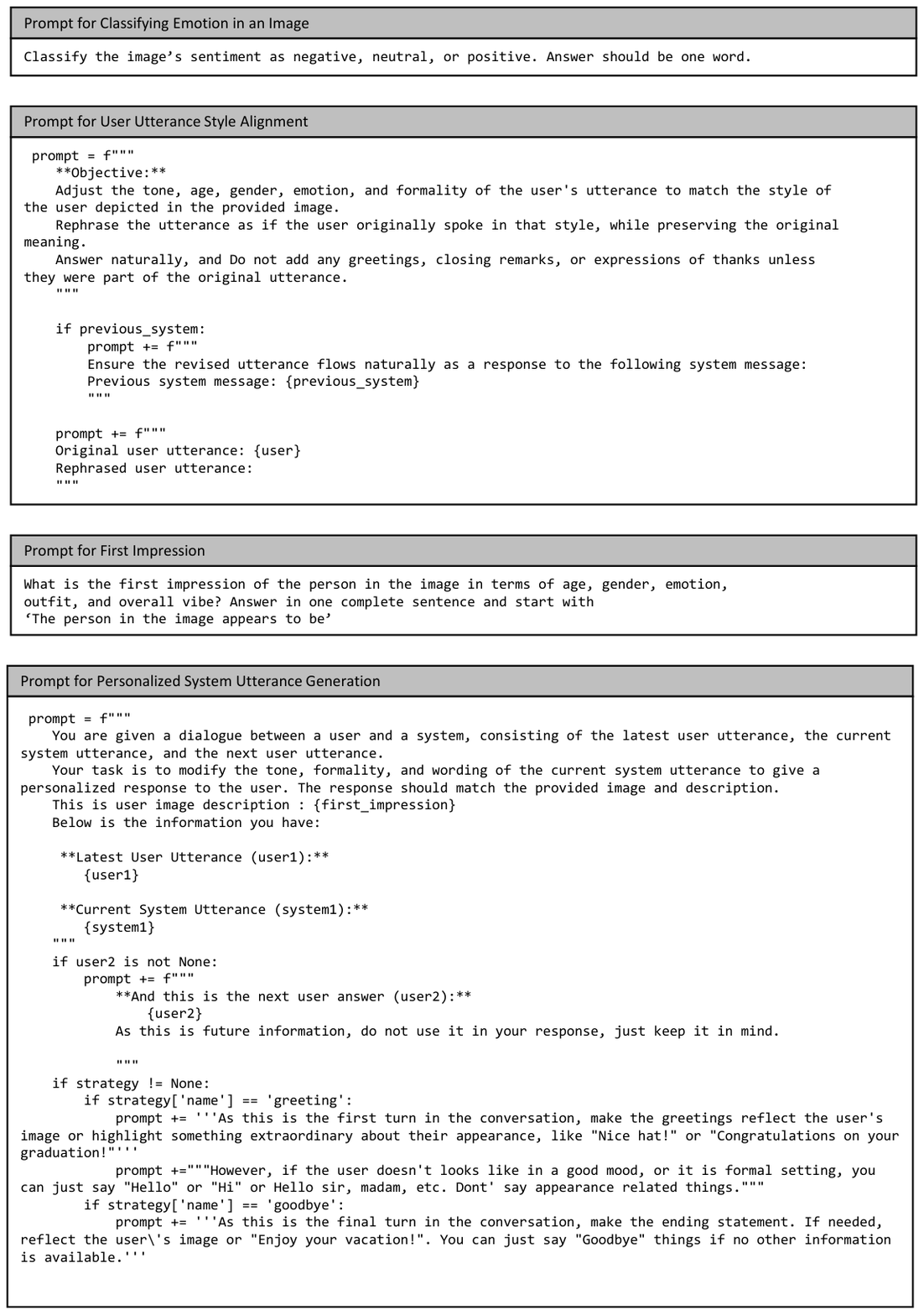}
\end{figure*}
\begin{figure*}[ht]
    \centering
    \includegraphics[width=\textwidth]{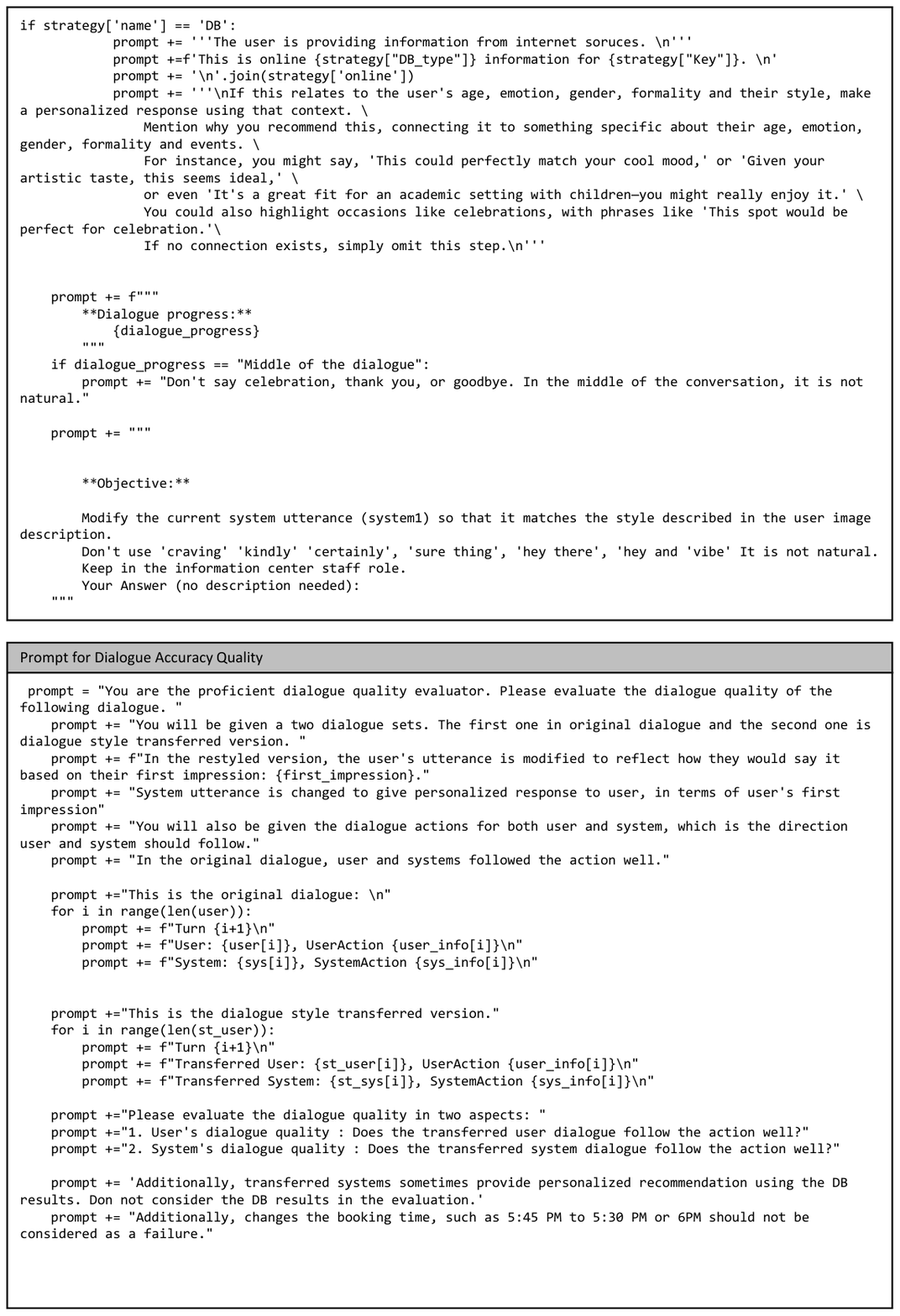}
\end{figure*}
\begin{figure*}[ht]
    \centering
    \includegraphics[width=\textwidth]{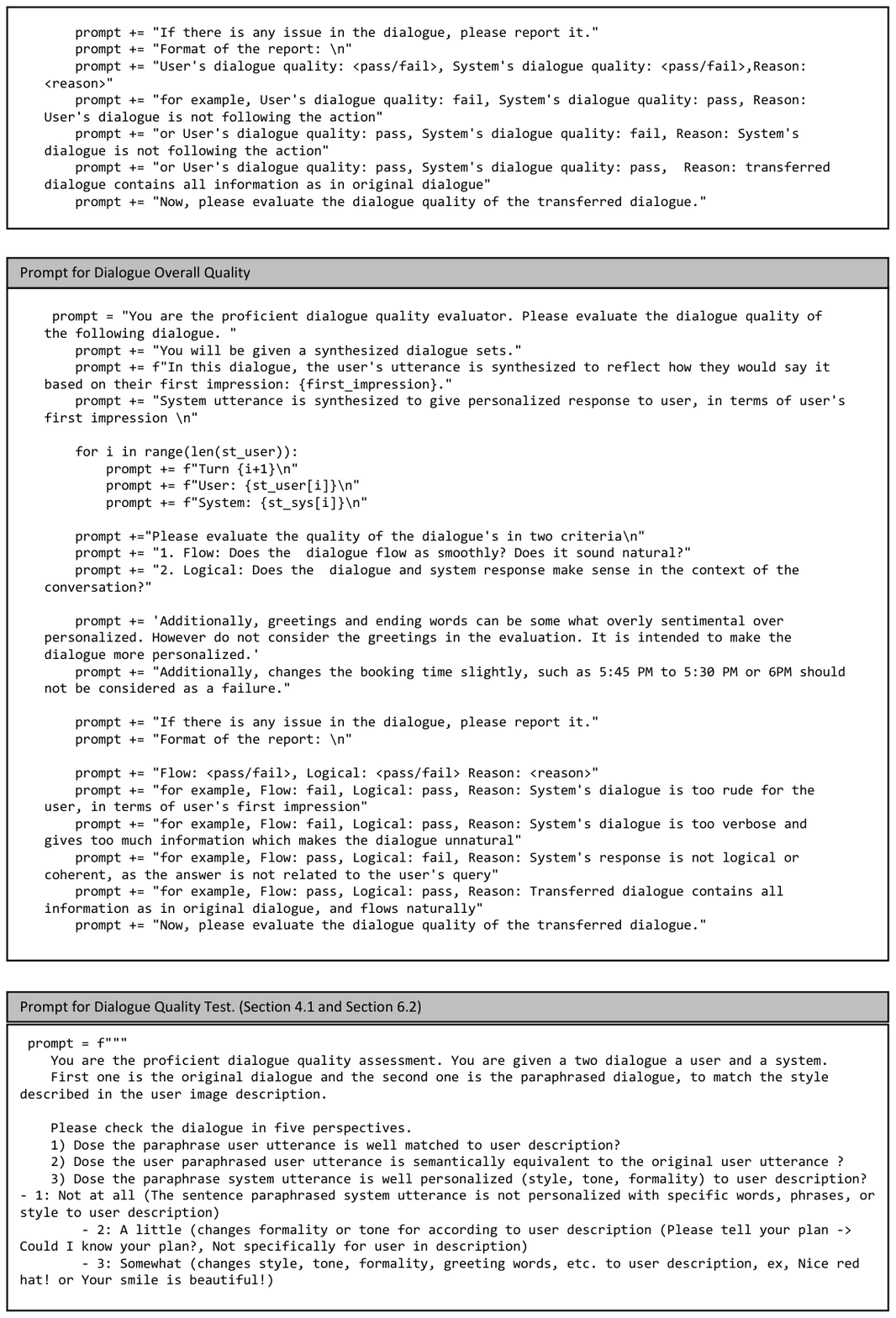}
\end{figure*}
\begin{figure*}[ht]
    \centering
    \includegraphics[width=\textwidth]{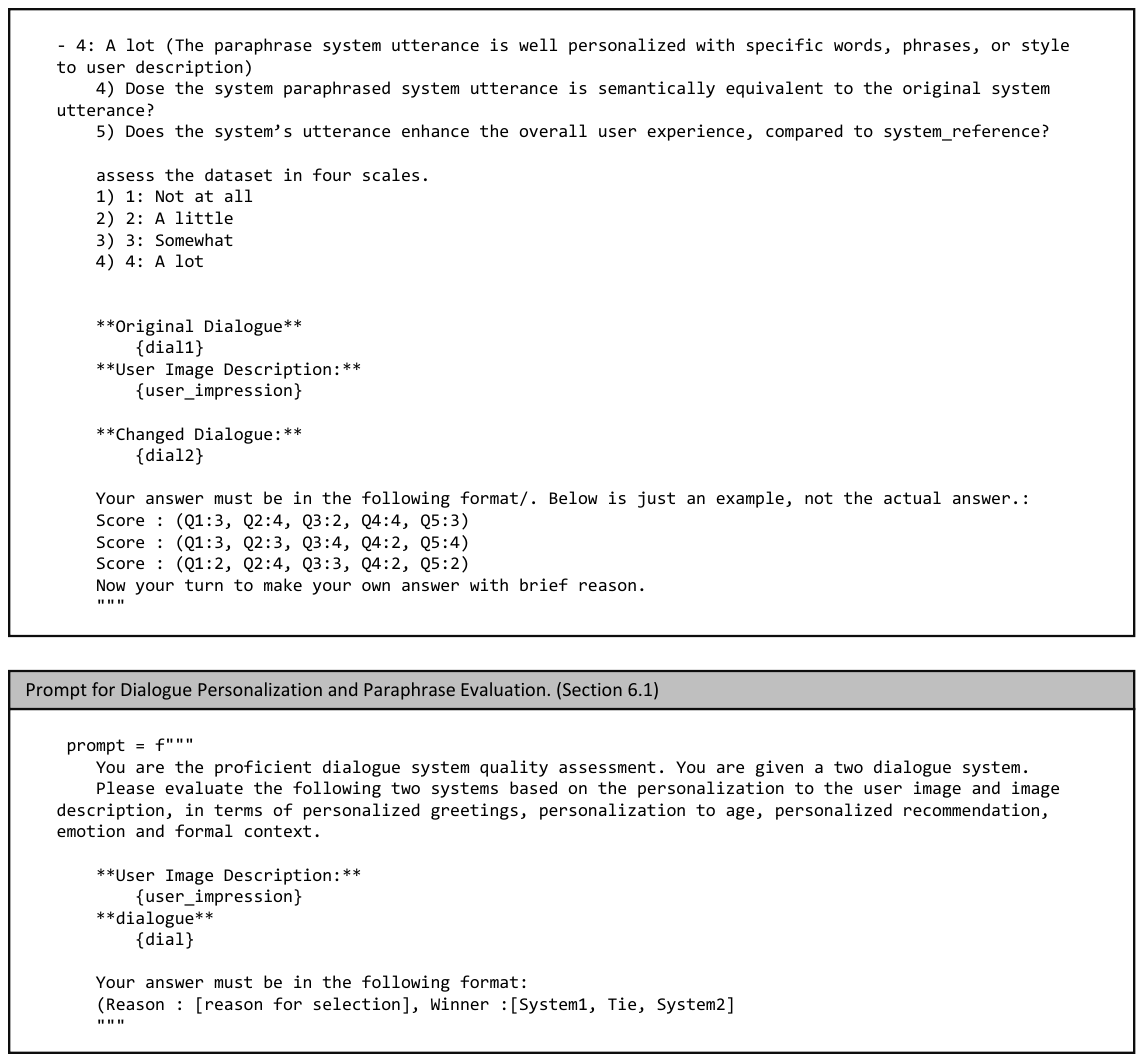}
    \caption{Promps templates.}
\label{fig:prompts}
\end{figure*}

% This licensing framework facilitates the adaptation of utterances from MultiWoZ 2.2 by incorporating images from the FFHQ dataset.

\end{document}